\documentclass[11pt]{article}
\usepackage[T1]{fontenc}
\usepackage[utf8]{inputenc}
\usepackage[a4paper, total={6in, 8in}]{geometry}
\usepackage{courier}

\title{{\bf A Comparison of Metric Learning Loss Functions \\for End-To-End Speaker Verification}}
\author{Juan M. Coria$^*$
        \and
        Herv\'{e} Bredin$^*$
        \and
        Sahar Ghannay$^*$
        \and
        Sophie Rosset$^*$}
\date{$^*$Universit\'{e} Paris-Saclay, CNRS, LIMSI, 91400, Orsay, France \\
      \vspace{0.5em}
      {\small\tt \{coria, bredin, ghannay, rosset\}@limsi.fr}}

\usepackage[numbers]{natbib}
\usepackage{graphicx, svg, amsmath, amssymb, multirow, tikz}
\usepackage[frozencache,cachedir=.]{minted}

\begin{document}

\maketitle

\begin{abstract}
    Despite the growing popularity of metric learning approaches, very little work has attempted to perform a fair comparison of these techniques for speaker verification. We try to fill this gap and compare several metric learning loss functions in a systematic manner on the VoxCeleb dataset. The first family of loss functions is derived from the cross entropy loss (usually used for supervised classification) and includes the congenerous cosine loss, the additive angular margin loss, and the center loss. The second family of loss functions focuses on the similarity between training samples and includes the contrastive loss and the triplet loss. We show that the additive angular margin loss function outperforms all other loss functions in the study, while learning more robust representations. Based on a combination of SincNet trainable features and the x-vector architecture, the network used in this paper brings us a step closer to a really-end-to-end speaker verification system, when combined with the additive angular margin loss, while still being competitive with the x-vector baseline. In the spirit of reproducible research, we also release open source Python code for reproducing our results, and share pretrained PyTorch models on \texttt{torch.hub} that can be used either directly or after fine-tuning.
\end{abstract}

\section{Introduction}

Given an utterance $x$ and a claimed identity $a$, speaker verification aims at deciding whether to accept or reject the identity claim.
It is a supervised binary classification task usually addressed by comparing the test utterance $x$ to the enrollment utterance $x_a$ pronounced by the speaker $a$ whose identity is claimed.
Speaker identification is the task of determining which speaker (from a predefined set of speakers $a \in S$) has uttered the sequence $x$.
It is a supervised multiclass classification task addressed by looking for the enrollment utterance $x_a$ the most similar to the test utterance $x$.
After a preliminary speech segmentation step, speaker diarization aims at grouping speech turns according to the identity of the speaker.

Whether we address speaker verification (this paper), speaker identification, or speaker diarization, the objective is to find a pair $(f,d)$ of representation function $f$ and comparison function $d$ with the following ideal property.
Given an utterance $x_a$ pronounced by a given speaker, any utterance $x_p$ pronounced by the same speaker should be closer to $x_a$ than any speech sequence $x_n$ uttered by a different one: $$d(f(x_a),f(x_p))< d(f(x_a),f(x_n))$$

State-of-the-art for speaker verification (like the \emph{x-vector} approach~\cite{Snyder2018XVectorsRD}) is no exception to the rule.
Their representation function $f(x) = n(h(x))$ is the composition of a handcrafted feature extraction step $h$ (e.g. filter banks or MFCCs) and a neural network $n$ trained for closed-set speaker recognition on a large collection of utterances from a large number of speakers.
Its comparison function $d$ is the composition of a trained Linear Discriminant Analysis (LDA) transform $l$, trained Probabilistic LDA (PLDA) scoring $p$~\cite{ioffe2016plda}, and adaptive s-norm score normalization $s$~\cite{matejka2017analysisNormalization}: $$d(f(x), f(x')) = s(p(l(f(x)), l(f(x')))$$

\begin{figure}[htb]
    \begin{center}
        \includegraphics[width=6cm]{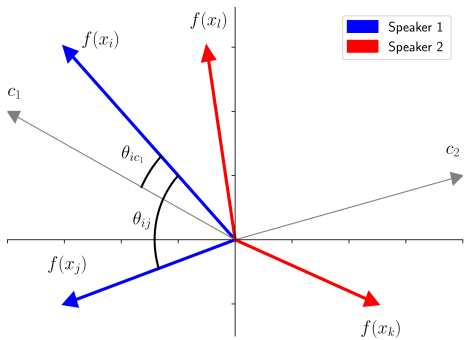}
    \end{center}
    \caption{Metric learning approaches aim at making representations of utterances of the same speaker close to each other, while separating utterances of different speakers as much as possible. In this example, the angular distance $\theta_{ij}$ between $f(x_i)$ and $f(x_j)$ should be small because utterances $x_i$ and $x_j$ were both pronounced by speaker \#1, while the distance between $f(x_l)$ and $f(x_i)$ should be large because $x_l$ is from speaker \#2. Some metric learning functions rely internally on centers $c_k$ that are trained jointly with the representation function $f$ and can be seen as a canonical representation of each speaker.}
    \label{fig:reprs}
\end{figure}

Metric learning techniques aim at simplifying the comparison function $d$ all the way down to the most simple distance function (e.g. euclidean or cosine distance), delegating all the hard work to the representation function $f$ (usually a trained neural network) that should ensure intra-class compactness and inter-class separability.
Figure~\ref{fig:reprs} depicts these concepts graphically in two dimensions.
Given training utterances $\{x_i\}$, the neural network~$f$ produces representations $\{f(x_i)\} \in \mathbb{R}^m$ such that the angular distance~$\theta$ between two utterances is small if they were pronounced by the same speaker (compactness) and large otherwise (separability).

A number of loss functions have been proposed to train such representation functions. While these approaches were mostly introduced for computer vision~\cite{hadsell_dimensionality_2006} and facial recognition in particular~\cite{schroff_facenet:_2015, liu_rethinking_2017, leibe_discriminative_2016, deng_arcface:_2018}, they have been rapidly adopted in other domains, like speaker verification~\cite{bredin_tristounet:_2017, chung_voxceleb2:_2018}, language identification~\cite{gelly_spoken_2017} and even natural language processing with sentence embedding~\cite{Reimers2019SentenceBERTSE}.

Speaker verification suffers from much of the same issues as face recognition, since utterances from a single speaker might differ in noise, phonetic content, mood, etc.
This is why we consider intra-class compactness and inter-class separability highly desirable properties for speaker embeddings as well.
Recent work in speaker verification~\cite{JHUVoxCelebSRC2019} has even shown that a simple cosine distance scoring with a metric learning loss can perform equally or better than a PLDA scoring on the same architecture.

Our first contribution is the systematic comparison of six metric learning loss functions for speaker verification, according to several criteria including raw performance, training time, and robustness. These losses can be loosely separated into two families (contrast-based and classification-based) and include the regular cross entropy loss, the additive angular margin loss~\cite{deng_arcface:_2018}, the center loss~\cite{leibe_discriminative_2016}, the congenerous cosine loss~\cite{liu_rethinking_2017}, the contrastive loss~\cite{hadsell_dimensionality_2006} and the triplet loss~\cite{schroff_facenet:_2015}. In particular, we show (like others did before us for face recognition~\cite{Srivastava2019comparison}) that the additive angular margin loss is better with respect to all considered criteria. More generally, margin-based loss functions (additive angular margin, contrastive, and to a lesser extent triplet loss) lead to representations that can be compared directly without heavy back end computations.

Based on this finding, our corollary second contribution is a step towards the definition of a truly (front) end to (back) end neural speaker verification approach. On the back end of the original x-vector approach, every one of LDA transform~$l$, PLDA scoring~$p$ and adaptive s-norm score normalization~$s$ needs its own (ideally disjoint) set of training data, making the approach quite complex and data-hungry.
This is why we define our architecture with only the cosine distance for scoring, and we show empirically that score normalization provides little to no significant improvement in our best models.

Existing end-to-end architectures in the literature often avoid PLDA as well~\cite{Zhang2017endtoendTriplet, Li2018angularEndtoend, Wan2018GE2E} but still rely on handcrafted features. Therefore, on the front end, we combine SincNet~\cite{ravanelli_speaker_2018} trainable feature extraction with the x-vector network architecture to build a fully trained representation function~$f$ that processes the waveform directly and does not rely on handcrafted features: $f(x) = h(n(x))$ becomes $f(x) = n(x)$. SincNet features have proven to outperform handcrafted features for some tasks~\cite{Pascual2019, Ravanelli2020} and have been used in conjunction with an angular margin loss in~\cite{Chagas2019AMSincNet} for speaker recognition on the simple TIMIT dataset.

Last but not least, a more practical third contribution is the joint release of the Python open-source code for training speaker embeddings with PyTorch and models pretrained on VoxCeleb speaker verification, which we argue is much simpler than the (\emph{de facto} standard) use of {\it Kaldi}~\cite{Povey_ASRU2011} for this purpose.

\section{Loss Functions}
\label{sec:losses}

This section defines the loss functions considered in the study and divides them in two families: the ones relying on classification with cross entropy loss, and the ones that rely on similarity between examples.

\subsection{Classification-based losses}
\label{ssec:softmax}

The first family of loss functions is derived from the cross entropy loss~$\mathcal{L}_\text{CE}$, initially introduced for multi-class classification:
\begin{equation}
\label{eq:softmax}
    \mathcal{L}_\text{CE} = - \frac{1}{N} \sum_{i=1}^{N} \log\left[ \frac{\exp(\sigma_{iy_i})}{\sum_{k=1}^K \exp(\sigma_{ik})} \right]
\end{equation}
where $N$ is the number of training examples (here, audio segments~$x_i$), $K$ the number of classes (here, speakers) in the training set, $y_i$ is the class of training sample $x_i$, and $\sigma_i$ is the output of a linear classification layer with weights~$C \in \mathbb{R}^{m \times K}$ and bias $b \in \mathbb{R}^K$:
\begin{equation}
   \sigma_i = f(x_i) \cdot C^T + b
\label{eq:linear}
\end{equation}
To facilitate the comparison with other metric learning loss functions, Equation~\ref{eq:linear} can be rewritten as follows:
\begin{equation}
   \forall k \;\; \sigma_{ik} = \left\|f(x_i)\right\| \cdot \left\|c_k\right\| \cdot \cos{\theta_{ic_k}} + b_k
\label{eq:class}
\end{equation}
where $\theta_{ic_k}$ is the angular distance between the representation~$f(x_i)$ of training sample~$x_i$, and $c_k$ the k$^{\text{th}}$ row of matrix~$C$.
Because the bias~$b$ is jointly trained with the representation function~$f$, the latter may learn to rely on the former to discriminate classes. A partial solution is to remove the bias:
\begin{equation}
    \label{eq:nobias}
    \forall k \;\; \sigma_{ik} = \left\|f(x_i)\right\| \cdot \left\|c_k\right\| \cdot \cos{\theta_{ic_k}}
\end{equation}
where the k$^{th}$ row of matrix $C$ can then be seen as a canonical representation of the k$^{th}$ speaker. 
Though removing the bias does improve performance slightly, this only partially solves the problem as the representation function~$f$ may still learn to encode speaker variability in their norm, which could in turn lead to a co-adaptation between representations and the classification layer. Hence, the congenerous cosine loss~\cite{liu_rethinking_2017} goes one step further by forcing the model to only rely on the angular distance between between $f(x_i)$ and $c_{k}$:
\begin{equation}
    \label{eq:coco}
    \forall k \;\; \sigma_{ik} = \alpha \cdot \cos{\theta_{ic_k}}
\end{equation}
where the hyper-parameter~$\alpha$ scales the cosine, further penalizing errors and favoring correct predictions.

None of the above loss functions address intra-class compactness specifically. The additive angular margin loss~\cite{deng_arcface:_2018} introduces a margin to penalize the angular distance between $f(x_i)$ and~$c_{y_i}$:
\begin{equation}
    \label{eq:arcface}
    \forall k \;\; \sigma_{ik} =
\begin{cases}
\alpha \cdot \cos(\theta_{ic_k} + m)& \text{if } y_i = k \\
    \alpha \cdot \cos \theta_{ic_k} & \text{otherwise}
\end{cases}
\end{equation}
where $m$ is the margin. This loss explicitly forces embeddings to be closer to their centers by artificially augmenting their distance by the margin.

The center loss~\cite{leibe_discriminative_2016} takes a different approach and adds a term to the cross entropy loss that penalizes the distance between training samples and a (jointly learned) representation $\gamma_k \in \mathbb{R}^m$ of their class $k$:
\begin{equation}
\label{eq:center}
    \mathcal{L} = \mathcal{L}_{\text{CE}} + \frac{\lambda}{2} \sum_{i=1}^{N} 1 - \cos \theta_{i{\gamma}_{y_i}}^2
\end{equation}

\subsection{Contrast-based Losses}
\label{ssec:contrastive}
While classification-based loss functions assume that the class of each training sample is known, this second family of loss functions relies solely on \emph{same/different} binary annotations: given a pair of training samples ($x_i$, $x_j$), the pair is said to be positive when $y_i = y_j$ and negative otherwise.

The contrastive loss~\cite{hadsell_dimensionality_2006} aims at making representations of positive pairs $\mathcal{P}$ closer to each other, while pushing negative pairs $\mathcal{N}$ further away than a positive margin $m \in \mathbb{R}^+$:
\begin{eqnarray}
\label{eq:contrastive}
\mathcal{L} & = & \sum_{(x_i, x_j) \in \mathcal{P}} (1 - \cos \theta_{ij})^2 + \nonumber \\
& & \sum_{(x_i,x_j) \in \mathcal{N}} \max(m - (1 -\cos \theta_{ij}), 0)^2
\end{eqnarray}
where $\theta_{ij}$ is the angular distance between $f(x_i)$ and $f(x_j)$.

The triplet loss~\cite{schroff_facenet:_2015} is defined in a similar way, but relies on triplets $(x_a, x_p, x_n) \in \mathcal{T}$, such that $y_a = y_p$ and $y_a \neq y_n$:
\begin{equation}
\label{eq:triplet}
\begin{aligned}
    \mathcal{L} = \sum_{(x_a, x_p, x_n) \in \mathcal{T}} \max(\cos \theta_{an} - \cos \theta_{ap} + m, 0)
\end{aligned}
\end{equation}
and aims at making the representation of positive samples $x_p$ closer to the anchor sample $x_a$ than the representation of any other negative samples $x_n$ by a positive margin $m \in \mathbb{R}^+$.

Because positive pairs~$\mathcal{P}$, negative pairs~$\mathcal{N}$ and triplets~$\mathcal{T}$ need to be sampled from the training set, they bring an additional computational cost that may slow down the training process. Morever, if tuples do not maximize the training signal, convergence issues may appear~\cite{schroff_facenet:_2015, hermans_defense_2017}. Those issues are usually addressed by selecting tuples carefully in a process known as \emph{mining}, making the whole process even more costly without any guarantee of the training stability.

To circumvent these issues, we use a slightly modified implementation of the triplet loss in our experiment:
\begin{equation}
\label{eq:triplet2}
\begin{aligned}
    \mathcal{L}_{T} = \sum_{(x_a, x_p, x_n) \in \mathcal{T}}\mathrm{sigmoid}(\alpha \cdot (\cos \theta_{an} - \cos \theta_{ap}))
\end{aligned}
\end{equation}
where $\alpha$ plays the same role as in Equations~\ref{eq:coco}~and~\ref{eq:arcface}. We hypothesized that the use of $\mathrm{sigmoid}$ may force all triplets to provide a normalized training signal, making large errors saturate to 1. Getting rid of the positive truncation also ensures that positive pairs keep getting closer and negatives pairs further apart, reducing the interest of keeping the positive margin.

While classification-based loss functions can only be used in a fully supervised setup, contrast-based loss functions that only rely on \emph{same/different} labels can be used in self-supervised scenarios~\cite{Pascual2019, Ravanelli2020} (e.g. by only using pairs with a high estimated probability of being positive or negative).

\section{Experiments}
\label{sec:experiments}

\subsection{Experimental protocol}
\label{ssec:protocol}

Experiments are conducted using VoxCeleb datasets~\cite{nagrani_voxceleb:_2017, chung_voxceleb2:_2018}, containing utterances in English. The whole VoxCeleb~2 development set (5994 speakers) serves as our training set. The VoxCeleb~1 development set (1211 speakers) is split into two parts: 41 speakers (whose name starts with U, V, or W) serve as our development set (1000 trials per speaker), the remaining 1170 speakers are used as cohort for adaptive s-norm score normalization. Final evaluation is performed on the official VoxCeleb~1 test set.

Each loss function comes with its own set of hyper-parameters, further described in Section~\ref{ssec:details}. Their optimal values are selected using grid search by training the model with each configuration for 20 hours and evaluating it on the development set in terms of equal error rate. The configuration leading to the best equal error rate is selected and used for further training the model for a total of 200 hours. Once training is completed, the best epoch is selected as the one leading to the best equal error rate on the development set. Finally, we apply the corresponding model on VoxCeleb~1 official test set (40 speakers) and report the equal error rate and corresponding $95\%$ confidence interval computed with the FEERCI package~\cite{feerci2018}. We also report the equal error rate after adaptive s-norm score normalization (whose cohort size is tuned on the development set). 
\subsection{End-to-end architecture}
\label{ssec:arch}

As stated in the introduction, the network architecture used in this set of experiments combines SincNet trainable feature extraction~\cite{ravanelli_speaker_2018} with the standard x-vector architecture~\cite{Snyder2018XVectorsRD} to build a fully end-to-end speaker verification system. Both SincNet and x-vector use the configuration proposed in the original papers (except for the SincConv layer of SincNet that uses a stride of 5 for efficiency).

\begin{figure*}[tb]
    \includegraphics[width=\textwidth]{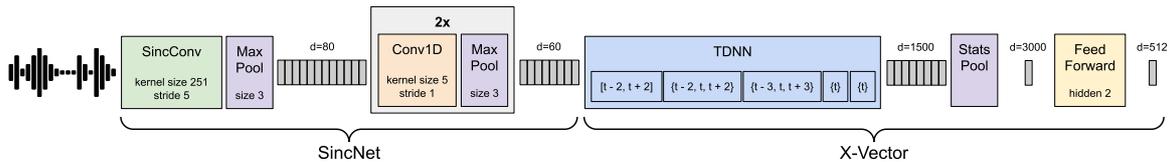}
    \caption{The end-to-end architecture used throughout the paper combines SincNet trainable features with the standard TDNN x-vector architecture.}
    \label{fig:sincarch}
\end{figure*}

As depicted in Figure~\ref{fig:sincarch}, the network takes the waveform as input and returns 512-dimensional speaker embedding. In practice, we use a 3s-long sliding window with a 100ms step to extract a sequence of speaker embeddings that are then averaged to obtain just one speaker embedding per file. These average speaker embeddings are then simply compared with the cosine distance.

\subsection{Implementation details}
\label{ssec:details}

All models were trained on GPU (NVIDIA Tesla V100) with Stochastic Gradient Descent using a fixed learning rate selected during the initial hyper-parameter grid search: we tried .001, .01, and .1. Mini-batches were built by stacking 3s~audio chunks extracted randomly from the training set, making sure each speaker was equally represented. Following lessons learned by others~\cite{mclaren_how_2018}, on-the-fly augmentation was used by dynamically adding random background noise from the MUSAN dataset~\cite{musan2015} with a random signal-to-noise ratio between 10 dB and 20 dB.

For classification-based loss functions, the batch size was fixed to 128 (from 128 different speakers). For contrast-based loss functions that expect pairs (or triplets) of training samples, a fixed number of audio chunks from a fixed number of different speakers were stacked to build mini-batches, before forming all possible pairs (or triplets) out of it. Both numbers were added to the set of hyper-parameters for these loss functions: we tried 20 and 40 for the number of speakers per batch, 2 and 3 for the number of audio chunks per speaker. The best hyper-parameter configurations found during the initial grid search are summarized in Table~\ref{tab:gridres}.

\begin{table}[h!]
    \centering
    \begin{tabular}{l|c|l}
        Loss function & LR & Hyper-parameters \\
        \hline
        \hline
        Cross entropy loss           & $10^{-1}$ & \\
        \hline
        Congenerous cosine loss      & $10^{-1}$ & $\alpha = 10$ \\
        \hline
        Additive angular margin loss & $10^{-2}$ & $\alpha = 10$ \\
                                     &           & $m = 0.05$ \\
        \hline
        Center loss                  & $10^{-1}$ & $\lambda = 1$ \\
        \hline
        Contrastive loss             & $10^{-1}$ & $m = 0.2$ \\
                                     &           & $3$ chunks $\times$ $20$ speakers \\
        \hline
        Triplet loss                 & $10^{-2}$ & $\alpha = 10^*$\\
                                     &           & $3$ chunks $\times$ $40$ speakers \\
    \end{tabular}
    \caption{Optimal hyper-parameters. LR stands for ``learning rate''. Hyper-parameters marked with~$^*$ were not tuned.}
    \label{tab:gridres}
\end{table}

\section{Results}
\label{sec:results}

\begin{figure}[htb]
    \centering
    \begin{tikzpicture}
        \node at (0,0) {\includegraphics[width=8.5cm]{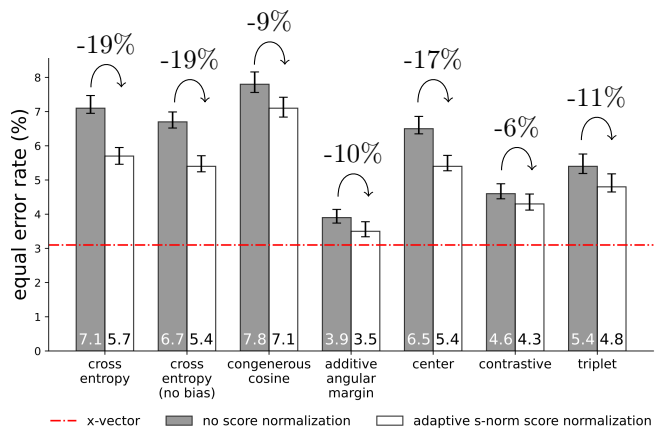}};
        \draw[->] (-3.2,2.1) .. controls +(0,0.4) and +(0,0.4).. node[above] {\small -19\%} (-2.8,2.1);
        \draw[->] (-2.1,1.9) .. controls +(0,0.4) and +(0,0.4).. node[above] {\small -19\%} (-1.7,1.9);
        \draw[->] (-1.05,2.4) .. controls +(0,0.4) and +(0,0.4).. node[above] {\small -9\%} (-0.65,2.4);
        \draw[->] (0.05,0.65) .. controls +(0,0.4) and +(0,0.4).. node[above] {\small -10\%} (0.45,0.65);
        \draw[->] (1.1,1.9) .. controls +(0,0.4) and +(0,0.4).. node[above] {\small -17\%} (1.5,1.9);
        \draw[->] (2.2,1) .. controls +(0,0.4) and +(0,0.4).. node[above] {\small -6\%} (2.6,1);
        \draw[->] (3.25,1.4) .. controls +(0,0.4) and +(0,0.4).. node[above] {\small -11\%} (3.65,1.4);
    \end{tikzpicture}
    \caption{Equal error rate on VoxCeleb~1 official test set, with or without adaptive s-norm score normalization. Relative improvement brought by score normalization is reported with curved arrows at the top. $95\%$ confidence intervals are depicted as vertical error lines. Performance of x-vector baseline~\cite{Snyder2018XVectorsRD} is reported for reference.}
    \label{fig:norm}
\end{figure}

\noindent \textit{Raw performance --} Figure~\ref{fig:norm} summarizes the raw performance of the proposed end-to-end architecture when trained with each loss function. We report the equal error rate on VoxCeleb~1 test set. The provided $95\%$ confidence intervals show that additive angular margin loss significantly outperforms all other loss functions. When combined with adaptive s-norm score normalization, it even achieves competitive performance with respect to the x-vector baseline that relies on handcrafted features (and for which we could not compute confidence intervals without access to the system output).

\noindent \textit{Robustness --} A closer look at the relative improvement brought by the score normalization step shows that additive angular margin loss and contrastive loss are the only ones for which the difference is not statistically significant. This suggests that the use of a margin leads to representations that are both better (in terms of raw performance) and more robust to domain mismatch.

\noindent \textit{Training time --} Despite training the models for 200 hours each, some of them were still improving on the development set when the time limit was reached: the congenerous cosine loss and the contrast-based (contrastive and triplet) losses. While the relative slowness of the latter can be explained by the fact that they both rely on on-the-fly tuple mining, we are still unsure as to why the former is so slow.

\section{Conclusion}
\label{sec:conclusion}

Overall, no matter the comparison criterion (raw performance, robustness, or training time), the additive angular margin loss is always better than the other loss functions that were considered in this study. If we really had to find one drawback, this would be the fact that it can only be used in a fully supervised learning scenario (contrary to its runner up, the contrastive loss).

As announced in the introduction, a model pretrained on VoxCeleb 2 with additive angular margin loss is available on {\small \texttt{torch.hub}}. Comparing two utterances for speaker verification is achieved with a few lines of commented Python code:

\vspace{0.5em}
\begin{figure}[h!]
    \centering
    \begin{minipage}{0.75\textwidth}
    \begin{minted}[xleftmargin=18pt,linenos,numbersep=2pt,fontsize=\footnotesize]{python}
    # load pretrained model from torch.hub
    import torch
    model = torch.hub.load('pyannote/pyannote-audio', 'emb')
    
    # extract embeddings for the whole files
    emb1 = model({'audio': '/path/to/file1.wav'})
    emb2 = model({'audio': '/path/to/file2.wav'})
    
    # compute distance between embeddings
    from scipy.spatial.distance import cdist
    import numpy as np
    distance = np.mean(cdist(emb1, emb2, metric='cosine'))
    \end{minted}
    \end{minipage}
\end{figure}

\section*{Reproducible research}
The companion Github repository\footnote{\texttt{github.com/juanmc2005/SpeakerEmbeddingLossComparison}} provides instructions to reproduce the main findings of this comparison.
It is based on the {\small \texttt{pyannote.audio}}~\cite{Bredin2020} toolkit that can be used to train (or fine-tune pretrained) models on a different dataset.

\section*{Acknowledgements}

This work was supported by the French National Research Agency (ANR) via the funding of the PLUMCOT project (ANR-16-CE92-0025) and was granted access to the HPC resources of IDRIS under the allocation 2019-AD011011182 made by GENCI.

\newpage
\bibliographystyle{plain}
\bibliography{references}

\begin{thebibliography}{10}

\bibitem{bredin_tristounet:_2017}
Herve Bredin.
\newblock {TristouNet}: {Triplet} loss for speaker turn embedding.
\newblock In {\em 2017 {IEEE} {International} {Conference} on {Acoustics},
  {Speech} and {Signal} {Processing} ({ICASSP})}, pages 5430--5434, New
  Orleans, LA, 2017. IEEE.

\bibitem{Bredin2020}
Herv{\'e} {Bredin}, Ruiqing {Yin}, Juan~Manuel {Coria}, Gregory {Gelly}, Pavel
  {Korshunov}, Marvin {Lavechin}, Diego {Fustes}, Hadrien {Titeux}, Wassim
  {Bouaziz}, and Marie-Philippe {Gill}.
\newblock {pyannote.audio: neural building blocks for speaker diarization}.
\newblock In {\em 2020 IEEE International Conference on Acoustics, Speech, and
  Signal Processing (ICASSP)}, Barcelona, Spain, May 2020.

\bibitem{Chagas2019AMSincNet}
J.~A. {Chagas Nunes}, D.~{Macêdo}, and C.~{Zanchettin}.
\newblock {Additive Margin SincNet} for {Speaker Recognition}.
\newblock In {\em 2019 International Joint Conference on Neural Networks
  (IJCNN)}, pages 1--5, July 2019.

\bibitem{chung_voxceleb2:_2018}
Joon~Son Chung, Arsha Nagrani, and Andrew Zisserman.
\newblock {VoxCeleb}2: {Deep} {Speaker} {Recognition}.
\newblock In {\em Interspeech}, pages 1086--1090. ISCA, 2018.

\bibitem{deng_arcface:_2018}
Jiankang Deng, Jia Guo, and Stefanos Zafeiriou.
\newblock {ArcFace}: {Additive Angular Margin Loss} for {Deep Face
  Recognition}.
\newblock {\em 2019 {IEEE} {Conference} on {Computer Vision} and {Pattern
  Recognition} ({CVPR})}, pages 4685--4694, 2019.

\bibitem{JHUVoxCelebSRC2019}
Daniel Garcia-Romero, Alan McCree, David Snyder, and Gregory Sell.
\newblock {JHU-HLTCOE system for VoxSRC 2019}, 2019.

\bibitem{gelly_spoken_2017}
G.~Gelly and J.L. Gauvain.
\newblock Spoken {Language} {Identification} {Using} {LSTM}-{Based} {Angular}
  {Proximity}.
\newblock In {\em Interspeech}, pages 2566--2570. ISCA, 2017.

\bibitem{feerci2018}
Erwin Haasnoot, Ali Khodabakhsh, Chris Zeinstra, Luuk Spreeuwers, and Raymond
  Veldhuis.
\newblock {FEERCI: A Package for Fast Non-Parametric Confidence Intervals for
  Equal Error Rates in Amortized O(m log n)}.
\newblock In Arslan Bromme, Andreas Uhl, Christoph Busch, Christian Rathgeb,
  and Antitza Dantcheva, editors, {\em 2018 International Conference of the
  Biometrics Special Interest Group, BIOSIG 2018}, International Conference of
  the Biometrics Special Interest Group (BIOSIG), United States, 2018. IEEE.

\bibitem{hadsell_dimensionality_2006}
R.~Hadsell, S.~Chopra, and Y.~LeCun.
\newblock Dimensionality {Reduction} by {Learning} an {Invariant} {Mapping}.
\newblock In {\em 2006 {IEEE} {Conference} on {Computer} {Vision} and {Pattern}
  {Recognition} ({CVPR})}, volume~2, pages 1735--1742, New York, NY, USA, 2006.
  IEEE.

\bibitem{hermans_defense_2017}
Alexander Hermans, Lucas Beyer, and Bastian Leibe.
\newblock In {Defense} of the {Triplet} {Loss} for {Person}
  {Re}-{Identification}.
\newblock {\em arXiv:1703.07737 [cs]}, March 2017.
\newblock arXiv: 1703.07737.

\bibitem{ioffe2016plda}
Sergey Ioffe.
\newblock {Probabilistic} {Linear} {Discriminant} {Analysis}.
\newblock In Ale{\v{s}} Leonardis, Horst Bischof, and Axel Pinz, editors, {\em
  Computer Vision – ECCV 2006}, pages 531--542, Berlin, Heidelberg, 2006.
  Springer Berlin Heidelberg.

\bibitem{Li2018angularEndtoend}
Y.~{Li}, F.~{Gao}, Z.~{Ou}, and J.~{Sun}.
\newblock {Angular Softmax Loss} for {End-to-end Speaker Verification}.
\newblock In {\em 2018 11th International Symposium on Chinese Spoken Language
  Processing (ISCSLP)}, pages 190--194, Nov 2018.

\bibitem{liu_rethinking_2017}
Yu~Liu, Hongyang Li, and Xiaogang Wang.
\newblock Rethinking {Feature} {Discrimination} and {Polymerization} for
  {Large}-scale {Recognition}.
\newblock {\em arXiv:1710.00870 [cs]}, October 2017.
\newblock arXiv: 1710.00870.

\bibitem{matejka2017analysisNormalization}
Pavel Matejka, Ondrej Novotn{\`y}, Oldrich Plchot, Lukas Burget, Mireia~Diez
  S{\'a}nchez, and Jan Cernock{\`y}.
\newblock {Analysis} of {Score} {Normalization} in {Multilingual Speaker
  Recognition.}
\newblock In {\em Interspeech}, pages 1567--1571, 2017.

\bibitem{mclaren_how_2018}
Mitchell Mclaren, Diego Castán, Mahesh~Kumar Nandwana, Luciana Ferrer, and
  Emre Yilmaz.
\newblock How to train your speaker embeddings extractor.
\newblock In {\em Odyssey 2018 {The} {Speaker} and {Language} {Recognition}
  {Workshop}}, pages 327--334. ISCA, June 2018.

\bibitem{nagrani_voxceleb:_2017}
Arsha Nagrani, Joon~Son Chung, and Andrew Zisserman.
\newblock {VoxCeleb}: {A} {Large}-{Scale} {Speaker} {Identification} {Dataset}.
\newblock In {\em Interspeech}, pages 2616--2620. ISCA, 2017.

\bibitem{Pascual2019}
Santiago Pascual, Mirco Ravanelli, Joan Serrà, Antonio Bonafonte, and Yoshua
  Bengio.
\newblock {Learning Problem-Agnostic Speech Representations from Multiple
  Self-Supervised Tasks}.
\newblock In {\em Interspeech}, pages 161--165, 2019.

\bibitem{Povey_ASRU2011}
Daniel Povey, Arnab Ghoshal, Gilles Boulianne, Lukas Burget, Ondrej Glembek,
  Nagendra Goel, Mirko Hannemann, Petr Motlicek, Yanmin Qian, Petr Schwarz, Jan
  Silovsky, Georg Stemmer, and Karel Vesely.
\newblock {The Kaldi Speech Recognition Toolkit}.
\newblock In {\em IEEE 2011 Workshop on Automatic Speech Recognition and
  Understanding}. IEEE Signal Processing Society, December 2011.
\newblock {IEEE Catalog No.: CFP11SRW-USB}.

\bibitem{ravanelli_speaker_2018}
Mirco Ravanelli and Yoshua Bengio.
\newblock {Speaker Recognition} from {Raw Waveform} with {SincNet}.
\newblock {\em 2018 IEEE Spoken Language Technology Workshop (SLT)}, pages
  1021--1028, 2018.

\bibitem{Ravanelli2020}
Mirco Ravanelli, Jianyuan Zhong, Santiago Pascual, Pawel Swietojanski, Joao
  Monteiro, Jan Trmal, and Yoshua Bengio.
\newblock {Multi-task self-supervised learning for Robust Speech Recognition}.
\newblock {\em ArXiv:2001.09239}, 2020.

\bibitem{Reimers2019SentenceBERTSE}
Nils Reimers and Iryna Gurevych.
\newblock {Sentence-BERT}: {Sentence Embeddings} using {Siamese BERT-Networks}.
\newblock In {\em EMNLP/IJCNLP}, 2019.

\bibitem{schroff_facenet:_2015}
Florian Schroff, Dmitry Kalenichenko, and James Philbin.
\newblock {FaceNet}: {A} {Unified} {Embedding} for {Face} {Recognition} and
  {Clustering}.
\newblock {\em 2015 {IEEE} {Conference} on {Computer} {Vision} and {Pattern}
  {Recognition} ({CVPR})}, pages 815--823, June 2015.
\newblock arXiv: 1503.03832.

\bibitem{musan2015}
David Snyder, Guoguo Chen, and Daniel Povey.
\newblock {MUSAN}: {A} {M}usic, {S}peech, and {N}oise {C}orpus, 2015.
\newblock arXiv:1510.08484v1.

\bibitem{Snyder2018XVectorsRD}
David Snyder, Daniel Garcia-Romero, Gregory Sell, Daniel Povey, and Sanjeev
  Khudanpur.
\newblock {X-Vectors}: {Robust} {DNN} {Embeddings} for {Speaker Recognition}.
\newblock {\em 2018 IEEE International Conference on Acoustics, Speech and
  Signal Processing (ICASSP)}, pages 5329--5333, 2018.

\bibitem{Srivastava2019comparison}
Yash Srivastava, Vaishnav Murali, and Shiv~Ram Dubey.
\newblock {A Performance Comparison} of {Loss Functions} for {Deep Face
  Recognition}.
\newblock {\em ArXiv}, abs/1901.05903, 2019.

\bibitem{Wan2018GE2E}
L.~{Wan}, Q.~{Wang}, A.~{Papir}, and I.~L. {Moreno}.
\newblock {Generalized End-to-End Loss} for {Speaker Verification}.
\newblock In {\em 2018 IEEE International Conference on Acoustics, Speech and
  Signal Processing (ICASSP)}, pages 4879--4883, April 2018.

\bibitem{leibe_discriminative_2016}
Yandong Wen, Kaipeng Zhang, Zhifeng Li, and Yu~Qiao.
\newblock A {Discriminative} {Feature} {Learning} {Approach} for {Deep} {Face}
  {Recognition}.
\newblock In Bastian Leibe, Jiri Matas, Nicu Sebe, and Max Welling, editors,
  {\em Computer {Vision} – {ECCV} 2016}, volume 9911, pages 499--515.
  Springer International Publishing, Cham, 2016.

\bibitem{Zhang2017endtoendTriplet}
Chunlei Zhang and Kazuhito Koishida.
\newblock {End-to-End Text-Independent Speaker Verification} with {Triplet
  Loss} on {Short Utterances}.
\newblock In {\em Interspeech}, pages 1487--1491, 2017.

\end{thebibliography}
\end{document}